\documentclass[10pt,twocolumn,letterpaper]{article}

\usepackage{verbatim}
\usepackage{cvpr}
\usepackage{times}
\usepackage{epsfig}
\usepackage{graphicx}
\usepackage{amsmath, amsthm, amssymb}
\usepackage{mdwlist}
\usepackage{mdframed}

\usepackage[ruled,vlined]{algorithm2e}
\usepackage[pagebackref=true,breaklinks=true,letterpaper=true,colorlinks,bookmarks=false]{hyperref}

\theoremstyle{definition}		% use "definition-style" font for the rest

\newcommand{\squishlist}{
 \begin{list}{$\bullet$}
  { \setlength{\itemsep}{0pt}
     \setlength{\parsep}{3pt}
     \setlength{\topsep}{3pt}
     \setlength{\partopsep}{0pt}
     \setlength{\leftmargin}{1.5em}
     \setlength{\labelwidth}{1em}
     \setlength{\labelsep}{0.5em} } }

\newcommand{\squishend}{
  \end{list}  }

\cvprfinalcopy % *** Uncomment this line for the final submission

\def\cvprPaperID{XXX} % *** Enter the CVPR Paper ID here

\begin{document}

%%%%%%%%% TITLE
\title{Out the Window:  A Crowd-Sourced Dataset for \\Activity Classification in Security Video}

\author{ 
Greg Casta\~{n}\'{o}n, Nathan Shnidman, Tim Anderson and Jeffrey Byrne \\
Systems \& Technology Research, Woburn MA \\
}

\maketitle
% \thispagestyle{empty}

%%%%%%%%% ABSTRACT
\begin{abstract}
%Activity detection in security video presents many unique challenges that are compounded by a lack of data.  Unlike datasets collected from social media sources, videos from security cameras have no guarantee that the relevant is in the video foreground.  Combined with the diverse range of viewing angles, near and far-field activities and paucity of reliable context clues, it can often be challenging to gather enough data to characterize this rich activity space using deep neural networks.  
The Out the Window (OTW) dataset is a crowdsourced activity dataset containing 5,668 instances of 17 activities from the NIST Activities in Extended Video (ActEV) challenge.  These videos are crowdsourced from workers on the Amazon Mechanical Turk using a novel ``scenario acting'' strategy, which collects multiple instances of natural activities per scenario.  Turkers are instructed to lean their mobile device against an upper story window overlooking an outdoor space, walk outside to perform a scenario involving people, vehicles and objects, and finally upload the video to us for annotation.  Performance evaluation for activity classification on VIRAT Ground 2.0 \cite{oh2011large} shows that the OTW dataset provides an 8.3\% improvement in mean classification accuracy, and a 12.5\% improvement on the most challenging activities involving people with vehicles.  

\end{abstract}

\section{Introduction}
\label{sec:intro}
Multi-camera security networks for monitoring activities at commercial, government and private facilities have been deployed for decades.  However, only recently have innovations in computer vision and machine learning achieved activity detection and classification performance such that automated detection of activities could replace human video monitoring.  However, these advances have been achieved primarily from social video sources \cite{carreira2017quo,monfort2018moments,karpathy2014large,abu2016youtube} and have yet to achieve similar performance for security video. 
%due to the significant differences in dataset quality and collection methodology between the social and security video domains.

%OTW OVERVIEW GRAPHIC

\begin{figure}
\includegraphics[width=3.3in]{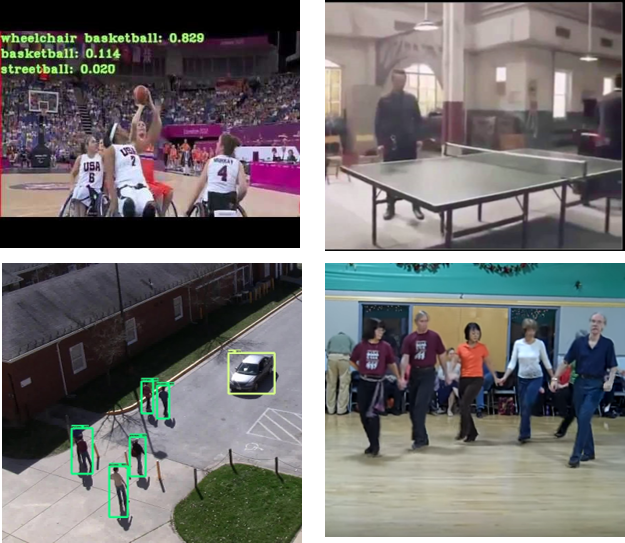}
\caption{Social media vs. security datasets for activity recognition.  Clockwise from top-left: Sports-1M, ActivityNet, Youtube-8M and VIRAT.  (lower left) security video datasets are dominated by viewpoints rarely seen in social video datasets, with background activities far from the camera.
}
\label{f:datasets}
\end{figure}

Two critical differences between social video and security video are the location of the subject within the frame and the perspective of the camera.  In social video, each video often contains a single activity, such that the foreground of the video is dominated by the activity or contextually-relevant visual information (e.g. a basketball court for the activity \textit{playing basketball}).  This form of activity framing is representative of a video journalism style, common on social media sites, which features the activity clearly in the foreground, shot from a first person point of view.  However, unlike social video the vast majority of security video involves infrequent activity with long idle periods.  When activities do happen, they can occur far away from the camera (as shown in figure \ref{f:datasets}) and only take up a small fraction of the spatial extent of a video frame.  Unlike activity detection with activity framing, which requires only temporal localization in untrimmed videos \cite{Ghanem18}, the key challenge with security video is that this requires both spatial and temporal localization.

%with an \textit{opening door} activity requires both spatial and temporal localization does not do much to suggest where the door is opened.

\begin{figure}
\includegraphics[width=3.3in]{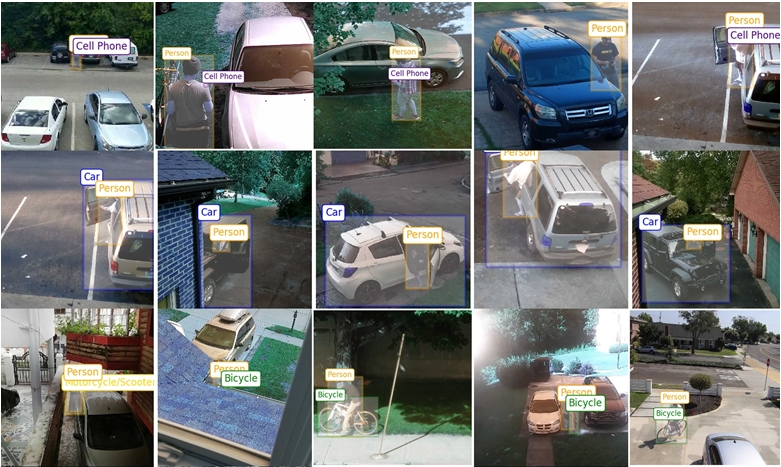}
\caption{Example frames from the Out the Window (OTW) dataset of vehicle and people activities and associated objects.}
\label{f:overview}
\end{figure}

Training an activity detection system for security video requires a deep training dataset with a diversity of viewpoints, actors, and scenes.   
%frequent obscuration of relevant activities, and issues with data collection.  
%While a great trove of videos with soft labels can be assembled from Youtube and tags \cite{abu2016youtube}, 
Collecting security video from varied viewpoints is challenging, since the scene may not contain the activities of interest.
%, and is further difficult to annotate.  
Furthermore, when it is collected, it is often from a very small number of scenes and actors \cite{oh2011large} due to the staffing limitations of the organization performing the data collection.  A dataset with a small number of viewpoints, actors and scenes will introduce a domain bias that limits the ability of the trained models to generalize to novel scenes.  
%and associated viewpoints may hinder the generalization performance of these models.  
Due to these challenges, no security dataset currently exists that is large enough to train an activity detection system from scratch.  Instead, recent approaches either fine-tune pretrained models on large social activity datasets using limited security data \cite{Gleason18} or 
incorporate domain knowledge into the activity representation to simplify optimization \cite{Zhang_2019_CVPR}.
However, training activity detection on a large, diverse training dataset specifically designed for the challenges of activity detection in security video would be preferred.

In this paper, we introduce a novel crowdsourcing approach to security data collection.  Crowdsourced workers from Amazon Mechanical turk are tasked to lean their cell phone against a window looking down onto your yard or driveway, and act out a prescribed scenario.  This scenario may be ``going to the grocery store'' or ''going on vacation'' which involve sets of activities such as carrying light and heavy objects (e.g. grocery bags, luggage) or loading or unloading vehicles (e.g. putting grocery bags into the trunk, putting luggage in back seat).  Turkers performed these scenarios, then the videos are uploaded for annotation.  
This data collection methodology addresses three key challenges:
\begin{itemize}
    \item {\em Scalable Data Collection.} Our system crowdsources the collection of videos of activities, allowing for video to be collected in parallel, fast, across the globe for a given set of activities.
    \item {\em Actor, Scene, Viewpoint and Label Diversity.} A dataset for general security should include a large variety of viewpoints, scenes, actors and activity classes.  Our approach naturally generates a wide variety of viewpoints and actors for each activity by leaving the precise viewpoint selection and actors up to the crowdsourced worker.  Furthermore, this approach enables collection of uncommon activity classes that are rarely found in social videos. 
    \item {\em Efficient Video Annotation.} Video annotation in untrimmed video is challenging due to the sparsity of rare activities, and the need to annotate every frame for ground truth.  Our approach includes weakly trimmed clips typically 1-2 minutes in which the turker performs the activity.  This weakly trimmed video is more efficiently annotated since it does not require search through a large video with no activities.  Furthermore, since the turker collects video in a relatively uncluttered scene, our system can leverage automataed object detection and tracking to ease annotator burden.   
\end{itemize}

\begin{figure}[t]
\includegraphics[width=3.4in]{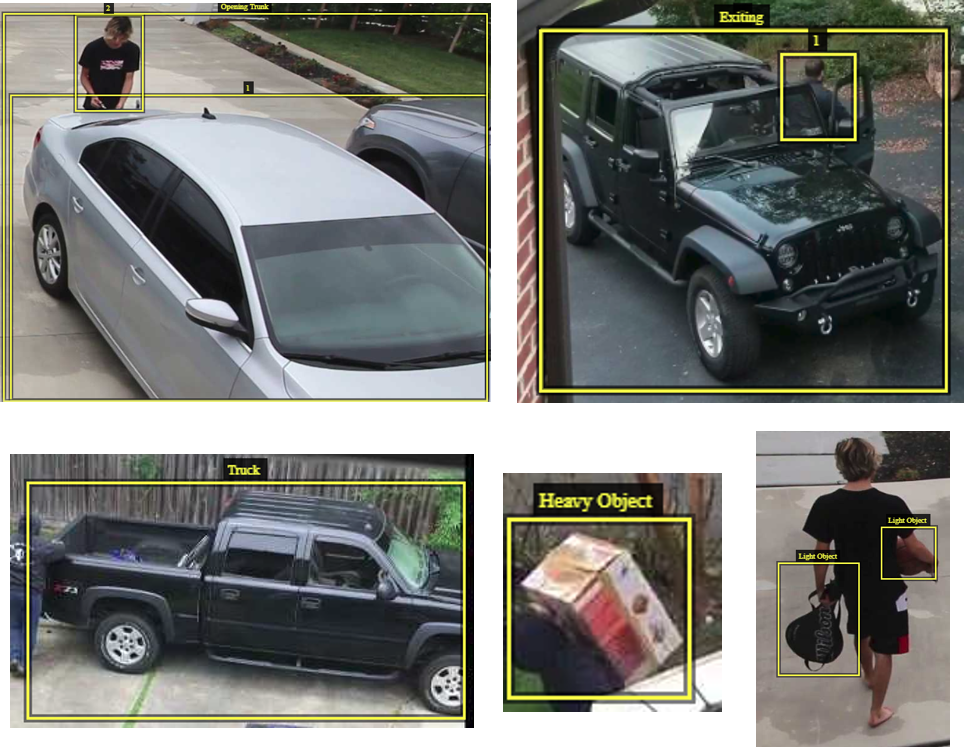}
\caption{OTW annotation examples for objects and activities.}
\label{f:annotation_examples}
\end{figure}

\begin{figure*}
\includegraphics[width=6.8in]{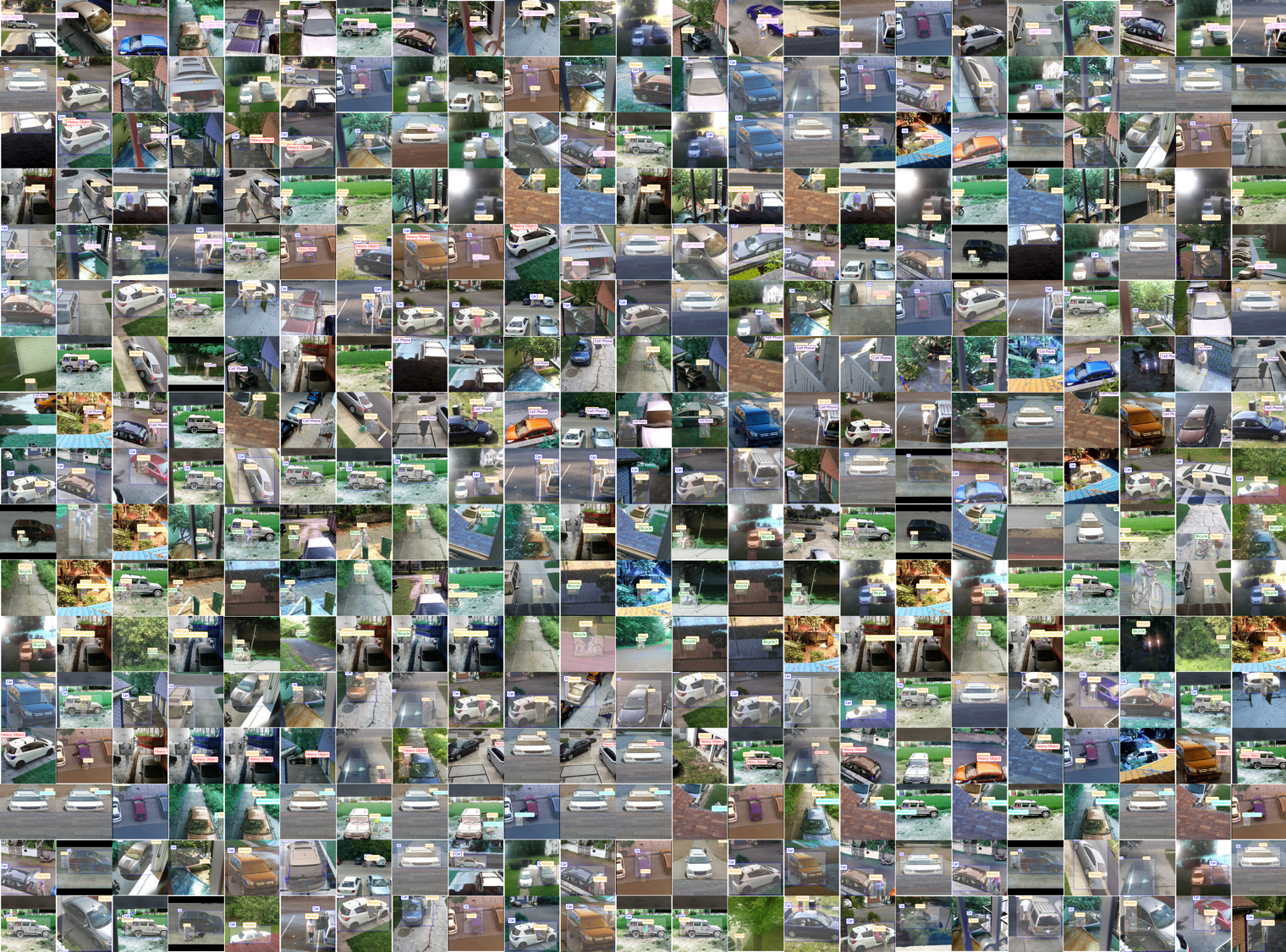}
\caption{Annotation examples from the OTW dataset, where each row corresponds to an activity class, each column corresponds to a single frame in a single instance of this activity and annotated boxes correspond to objects contributing to this activity.  Activities rowwise: Carrying (Small), Closing Trunk, Unloading Vehicle, Conversation, Loading Vehicle, Opening Door, Talking on Phone, Texting on Phone, Exiting, Dismounting Bike, Mounting Bike, Riding Bike, Entering, Carrying (Large), Pushing Cart, Opening Trunk, Closing Door.  For best results, view in color and zoom into the montage in the PDF. }
\label{f:otw_montage}
\end{figure*}

In this paper, we describe our approach to data collection, annotation and post-processing of the Out the Window (OTW) dataset. To the best of our knowledge, our approach is the first to apply crowdsourcing techniques to security video collection, introducing a scalability and diversity of actors, viewpoints and scenes to the problem of activity detection in security video.  This diversity is highlighted in the sample of OTW annotations shown in figure \ref{f:overview}, figure \ref{f:annotation_examples} and figure \ref{f:otw_montage}.  The 17 activity labels and 9 object labels collected along with collection statistics are shown in figure \ref{f:dataset_statistics}. We use the OTW dataset to train a baseline Temporal Segment Network \cite{Wang18} for activity classification, and compare performance with a baseline trained on the VIRAT dataset \cite{oh2011large} .  This demonstrates the utility of our method of data collection for activity classification on security video.   This dataset will be made publicly available under a Creative Commons Attribution 4.0 International (CC BY 4.0) license.  

\section{Related Work}
\label{sec:relatedwork}

\begin{figure*}
\centering
\includegraphics[width=6.85in]{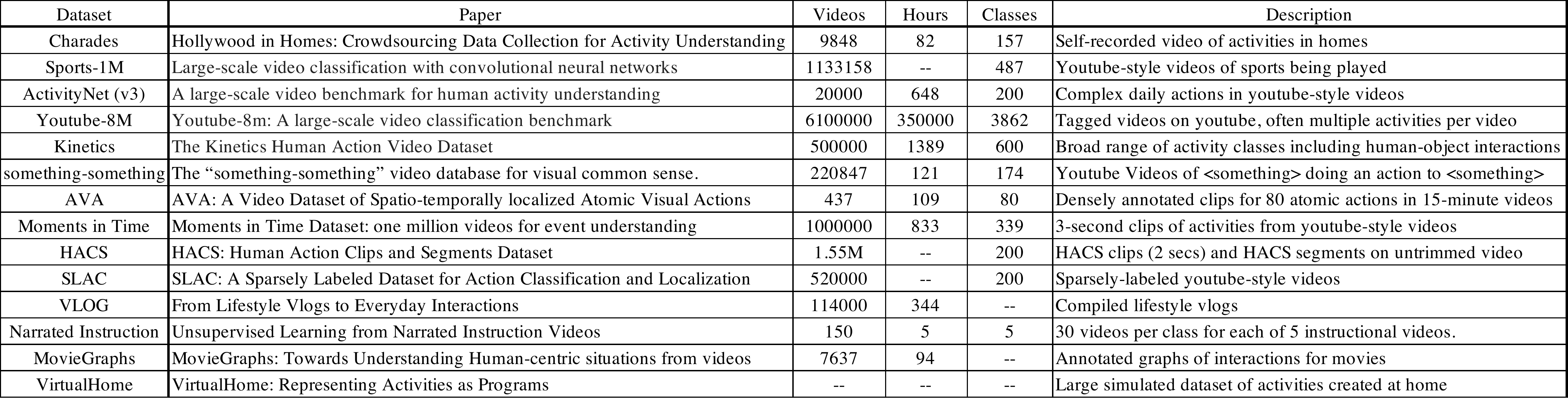}
\caption{A survey of publicly available, large scale datasets for activity recognition.}
\label{f:other_datasets}
\end{figure*}

Figure \ref{f:other_datasets} shows a summary of the available public datasets for activity detection.  Classic datasets, such as the KTH \cite{laptev2004recognizing} and Weizmann \cite{gorelick2007actions} datasets, employed actors or graduate students to perform activities in front of a fixed camera for activity classification.  These approaches established baselines for action classification, particular from motion, but lacked many of the aspects of natural video.  
Datasets like CAVIAR \cite{fisher2005caviar} and Casia \cite{zhang2007trajectory} followed, which involved labeled activities performed in outdoor scenes.  Approaches like the UCF-Aerial \cite{ucf_aerial}, VIRAT \cite{oh2011large} datasets continued to use this methodology - set up a camera either surveilling a public area like a parking lot, or hire actors to perform the activities in question.

In contrast, modern large scale datasets are dominated by videos from social media.  Classic datasets were sufficient to evaluate model-driven activity detection, however they contained insufficient examples to generalize to high-parameter deep networks.  The emergence of social video websites like Youtube with user-tagged video spurred the emergence of large datasets such as HMDB51 \cite{kuehne2011hmdb}, UCF101 \cite{soomro2012ucf101} and the associated THUMOS challenge \cite{THUMOS14}.  These datasets collected thousands of videos off social media sites to support activity classification.  The approach of collecting video from social sites is now common practice, and was used to create Youtube Sports-1M \cite{karpathy2014large}, ActivityNet \cite{caba2015activitynet}, Kinetics \cite{kay2017kinetics}, Atomic Visual Actions (AVA) \cite{gu2018ava} and Youtube-8M \cite{abu2016youtube} and Human Actions Clips and Segments (HACS) \cite{zhao2019hacs}.

However, the classes available by this style of collection are limited to searchable tags on Youtube.  Recently, researchers have turned to crowdsourcing using an on-demand workforce such as Amazon Mechnical Turk (AMT)
%\cite{mturk} 
to perform targeted collection of specific types of activities.  The Hollywood in Homes \cite{sigurdsson2016hollywood} dataset introduced this style of approach, randomly generated scripted tasks for humans to perform in their home.  The Something-Something \cite{Goyal17} dataset used a similar approach to collect examples of \textit{something} performing an action on \textit{something else}.  These crowdsourced strategies of video collection are closely related to our approach, however we believe that our collection methodology is the first application of crowdsourced data collection to security video. 

%While this is by no means a complete accounting of all the video datasets that have been collected, it illustrates the class of data collections that have been undertaken:  scripted individual activities, annotated acted activities, social video collect and crowdsource annotation for human activity recognition.  We believe that ours is the first work to apply crowdsourcing techniques to security video collection, introducing a scalability and diversity of actors, viewpoints and scenes heretofor unavailable to the problem of activity detection in security video.  

\section{Data Collection}
\label{sec:system}

The key challenges in crowdsourced data collection are scalability and diversity.  Our goal was to scale our data collection while collecting an equal amount of each activity performed by many actors, in many scenes, from many viewpoints.  Because of the lack of a centralized, Youtube-style repository with soft labels, no simple solution exists.  Scripted data collection, like classic activity datasets \cite{laptev2004recognizing,gorelick2007actions}, suffers from an issue of verisimilitude; actors who know they are supposed to perform a given task do not perform that task naturally.  Further, scripted datasets \cite{oh2011large} can only afford to pay so many actors and set up so many cameras, leaving them without the diversity of viewpoints, scenes and actors that we were attempting to achieve.  While publically available live-streams of outdoor areas such as Youtube Live and assorted webcam sites may present an appealing alternative for large-scale data collection, it can be difficult to localize activities of interest, as most of the time nothing is happening.  There are also assorted issues with terms of service.

The OTW dataset overcomes these issues by leveraging an asset unavailable the last time a large-scale security dataset was collected: the ubiquity of high quality cell-phone cameras, even in third world countries.  These phones frequently record far superior video to the average security camera at up to 60 FPS and 1920x1080 resolution, which allows any given person to act as their own source of security data.  

To gain access to these people and phones, we turned to the Amazon Mechanical Turk (AMT), which has shown to be an excellent source of crowdsourced labor.  Further, this labor tends to be relatively affordable, enabling scalable collection of large amounts of data for a modest sum.  There has been ample work about optimizing data acquisition and labeling on the Amazon Mechanical Turk.  Some emphasize pricing strategies to maximize collection while minimizing cost, while others detail \cite{vedantam2015cider} methods for achieving consensus to reduce label noise.  However, the focus of this collection effort was on quality of data received, not annotation efficacy or pricing, and so we did not explore that space.

\begin{figure*}
\centering
\includegraphics[width=6.85in]{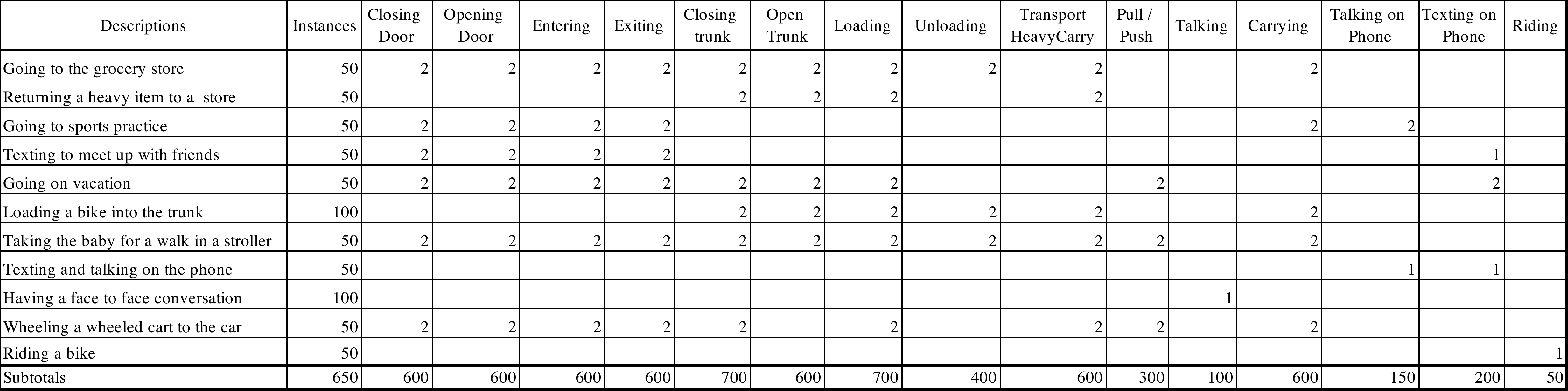}
\caption{``Scenario acting'' decomposes a complex activity (e.g. going to the grocery store) into a set of desired activities (e.g. closing door) to avoid actors generating unnatural activities, since they are not told explicitly what to do.   We chose 11 scenarios to cover the desired activities to hit a target of desired activity instances for collection.
}
\label{f:activity_counts}
\end{figure*}

\subsection{Approach}
\label{ss:approach}

Inspired by the Hollywood-in-Homes (Charades) dataset \cite{sigurdsson2016hollywood}, the Human Information Task (HIT) that we gave to each worker was as follows:  lean your cell phone against a window looking down onto your yard or driveway, and act out a prescribed scenario.  Each scenario we gave was a broad description of an activity they might do, such as load groceries into their car or have a conversation, with each scenario \textit{implicitly} causing the actor to perform certain actions (e.g. "opening a car door" or "enteirng a car").  While people act differently when they know they are on camera, we hypothesize that we can mitigate this by not cuing actor attention towards specific activities by letting the actors know which are important.  This approach also leads to diverse and often unanticipated examples of activities that should help generalization of deep models.

In order to collect data on the 17 activities of interest, we mapped them to 11 scenarios, shown in Table \ref{f:activity_counts}.  We built a simple UI which first provides a clear explanation of the task and requirements to prospective workers, then signs a consent form for the release of personally identifiable information (PII), and finally provides in-depth instructions and a way to upload the collected video to an Amazon S3 bucket (used again for scaling purposes).  In addition to the consent form, we advised workers to hide PII (e.g. their face, license plates), as none of this information is in scope for the dataset being collected.  Finally, all data collection was performed under review of an external institutional review board (IRB) to certify data collection protocols for human subjects research.

\subsection{Lessons Learned}
\label{ssec:lessons}

Our HIT differed in two primary ways from the common types of HITs available on the Amazon Mechanical Turk: it was significantly more complex and paid significantly better.  What we came to learn was that this represented a fundamental risk for Turkers; many high-paying tasks, including ours, are gated by the percentage of HITs that a worker has had approved.  If they fall below these common thresholds (often 95, 98 or 99\%), they have to go do a number of low-paying tasks in order to rise above these thresholds again.  Given that this metric does not account for length, payment, or difficulty of task, investing 15-20 minutes in a task that pays \$5 is a higher-risk move than doing a series of shorter tasks that pay worse.

As a result, we found that the driving factor in the rate at which people accepted our HITs was our reputation.  In the first week of release, we collected on the order of 50 videos; as reviewers on social networks like Turkopticon,
%\cite{turkopticon}, 
Reddit 
%\cite{redditmturk} 
and Turkerview 
%\cite{turkerview} 
confirmed that we were acting in good faith, this scaled quickly to hundreds of videos a week.

\section{Data Annotation}
\label{sec:annotation}
Creating affordable, high-quality annotations for video data collection has been a common challenge in computer vision research.  In object annotation for object detection in imagery, the challenge is to come to a consensus on the bounding box that correctly covers the object of interest in the image.  This problem is compounded in our case, as many activities occur during our videos, and the objects involved in these activities frequently move around.  The naive approach would be to annotate every activity in every video frame-by-frame.  Unfortunately, to annotate a corpus of over 11 hours of video, this would take about 206 days of annotation for a single annotator who could annotate a frame every five seconds, eight hours a day, every day.  
As a result, we decide to use a keyframe approach to annotation that combined sparse annotation with object detection and tracking to produce the rich annotations are expected in activity detection datasets.

\begin{figure*}
\centering
\includegraphics[width=6.9in]{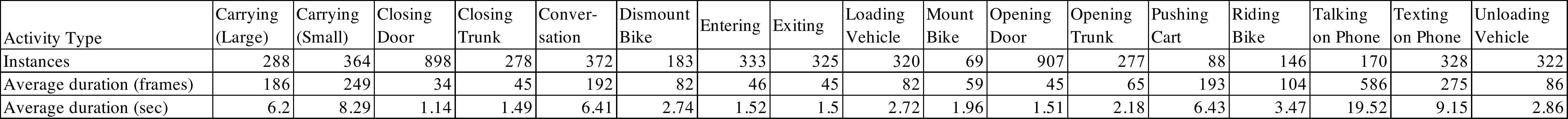}

\vspace{4pt}

\includegraphics[width=4in]{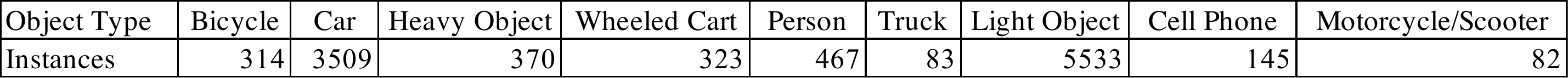}
\caption{OTW dataset statistics.  The 17 activity labels and number of instances with 8 object label and number of object instances.
}
\label{f:dataset_statistics}
\end{figure*}

Our data annotation process is as follows.  We begin by having an expert clearly define the beginning and end point of every activity of interest in the dataset.  As an example, an \textit{opening door} action begins when somebody's hand touches the door and ends when the door stops moving.  When possible, we tried to keep these definitions close to those in the VIRAT dataset \cite{oh2011large}.  We then create a project for each video using the VIA Annotation Tool \cite{dutta2016via} hosted on an Amazon S3 bucket, and assign each annotator a list of projects to annotate.  The analyst annotates only the first and last frame for each activity in a video, drawing bounding boxes around both the activity and any object participating in the activity (e.g. a car and a person for \textit{opening door}).  This approach dramatically reduces annotator burden - often requiring less than 20 frames in a video be annotated.

The key challenge with annotation by tracking is due to tracking error.  Linear interpolation may serve as a viable way for certain activities (e.g. \textit{opening door}, \textit{loading car}), other activities like \textit{carrying (light object)} involve a person moving, often not in a straight line.  We achieved this interpolation by detecting objects, tracking those objects, and corresponding likely object trajectories for the objects involved in the frame. Object detection was performed framewise using the Mask RCNN object detector \cite{He17}, fine-tuned to the security domain the VIRAT Ground 2.0 dataset.  Tracking was performed using the SORT tracker \cite{Bewley16} with a constant velocity, constant aspect ratio 7-state model:  $[x, y, a, r, \Dot{x}, \Dot{y},\Dot{a}]$.  Data association was achieved by the greedy approach of choosing the track with maximum spatial intersection-over-union (IoU) with the current state and following that track until its end; the state was initialized with the annotations.  The bounding box for the activity is defined as the convex union of the objects participating in it.

We address the risk of tracking error by identifying which frames are annotated by tracking and which frames are annotated by a human.  Relying on automated approaches to interpolation is always risky, since there are many sources of detection error, particularly when there is high potential for occlusion.  However, performing framewise detection on high-resolution cell phone camera videos mitigates the chances of track breakage.  Furthermore, the data collection methodology requires that scenes are sparse (e.g. containing at most two consented persons) and videos are weakly trimmed (e.g. activity occurs somewhere within a 1-2 minute video).  These two properties suggest that a tracking based annotation will have limited false alarms due to the relatively uncluttered scenes.  

\subsection{Dataset Statistics}
\label{ssec:statistics}

\begin{figure*}
\includegraphics[width=3.55in]{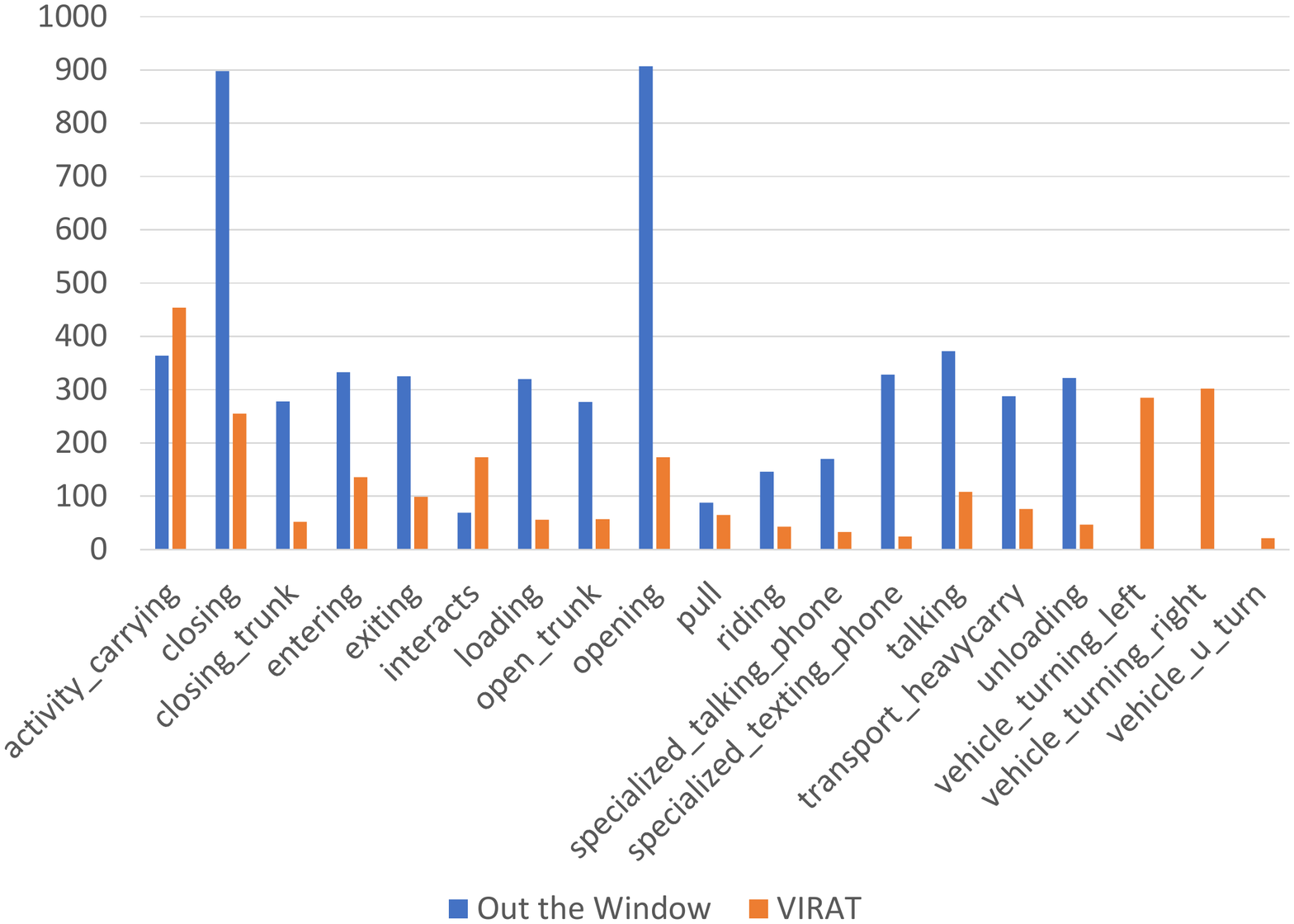}
\includegraphics[width=3.55in]{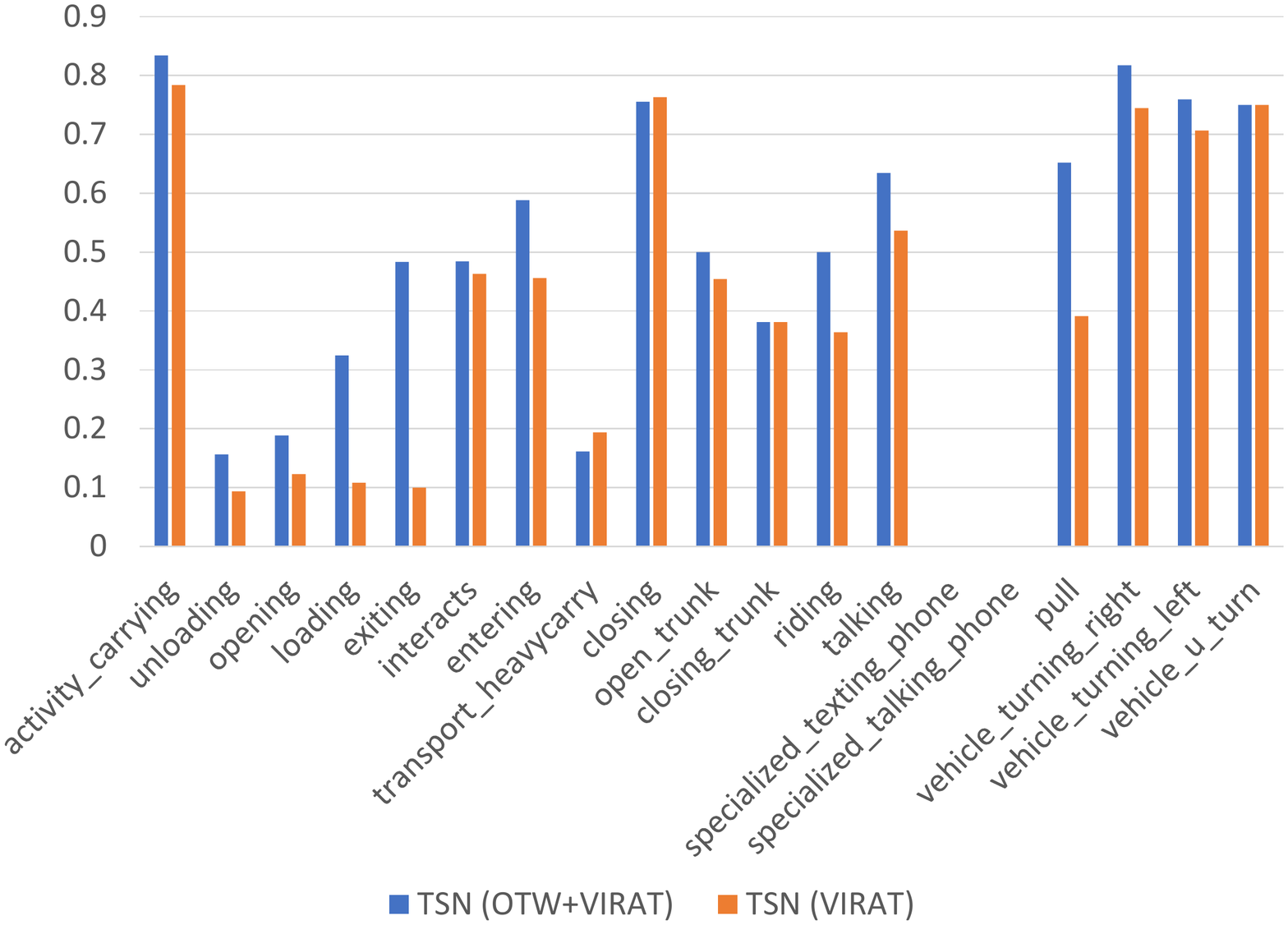}
\caption{ (Left) The OTW dataset size compared with the training/validation split of the VIRAT Ground 2.0 dataset (Validation + Training).  (right) Mean classification accuracy of a TSN trained using OTW+VIRAT-train vs. the performance of one trained on only VIRAT-train, evaluated on VIRAT-val.
}
\label{f:tsn_otwdiva_vs_diva}
\end{figure*}

Figure \ref{f:dataset_statistics} shows the overall Out the Window dataset statistics.  In total, the collection resulted in 11 hours of video containing 6.1 hours of activities.  This is a notably high ratio for security video collection, which generally features sparse activity collection.  We were able to achieve a large number of counts of vehicle-related activities like \textit{opening}, \textit{loading} and \textit{entering} due to having multiple scenes involving these activities.  Our long-form scenes tended also to result in more of the desired activities per video, since we found it difficult to combine activities like \textit{riding a bike} with other activities without being too explicit as to which activities should be performed.  We found also that many people just did not own a wheelbarrow or cart to push, limiting the number of \textit{pushing cart} activities that we could get.

Figure \ref{f:dataset_statistics} also shows that while we have less exemplars of certain activities like \textit{talking on the phone}, many of these activities have longer duration.  Despite having twice as many examples of texting as opposed to talking, we have nearly the same number of minutes of \textit{texting on phone} as \textit{talking on phone}.  This is likely due to people texting, doing something else, and then texting again, while they hold phone conversations for long periods of time.

\section{Performance Evaluation}
\label{sec:evaluation}
In order to evaluate the benefit of this data collection, we utilized one of the largest security video datasets available, the VIRAT Ground 2.0 \cite{oh2011large} dataset.  This dataset contains 11 scenes, each shot persistently from a single static security camera.  We used a subset of 5 scenes from this dataset which contained the majority of the activity instances.  This dataset is representative of a particular style of security data collection: it contains a small set of actors, scenes and viewpoints, and so models trained exclusively on this data often generalize poorly to novel scenes.

In order to evaluate whether or not OTW data had an impact a generic security problem, we trained a Temporal Segment Network (TSN) \cite{wang2016temporal} for the task of activity classification.  This baseline network was trained using a training/validation split of the VIRAT data, and compared its performance to the same networked trained on both the VIRAT training data and the OTW dataset.  We chose TSNs over other approaches that rely on a fixed-length input (e.g. \cite{simonyan2014two,wang2018non}) because it addresses the time dilation involved in many of these activities.  In the train and validation sets, the \textit{opening} activity is performed as fast as 18 frames (.6 seconds) and as slowly as 300 frames (10 seconds).  Temporal Segment Networks hypothesize that the first third (for $k=3$) of these activities will look similar.  

The  metric for this performance evaluation is mean classification accuracy for labeling a trimmed video clip with one of 19 activity labels.  This performance evaluation considers closed set classification, and evaluates only activity classification given localization on trimmed videos, as a baseline before considering activity detection on untrimmed videos.  

We initialized this network with a pre-trained model from the Kinetics dataset, and fine-tuned it using the default settings: 7 segments and a learning rate of .001.  We trained for 200 epochs, terminating when loss on the validation set plateaued.  Note that this approach is a bit unfair - we terminate based on validation loss and also evaluate on the validation set.  However, absent the ability to self-evaluate on the test set, it represents the most principled choice for the default split of the data.

\subsection{VIRAT Train/Validation Split}
\label{ssec:phase1}
Our training and validation split was on 5 scenes (each with a single camera) from the VIRAT dataset.  There are a total of 1404 activities in the Train set and 1203 activities in the Validation set.

The VIRAT dataset illustrates one of the most common issues with security video: highly imbalanced classes.  While there are hundreds of examples of people carrying objects, there are only 13 training examples of vehicles U-turning, and 8 examples in the Validation set.  This has the dual effect of making it exceedingly hard to train a detector or classifier for these rare activities, and not incentivizing a classifier to perform well on these activities.  Because the metrics reflect the total number of activities detected, a classifier that does perfectly on \textit{u-turn} will only pick up 8 detections, while one that does perfectly on \textit{activity carrying} will contribute 199.

Figure \ref{f:tsn_otwdiva_vs_diva} (left) shows the number of events collected in our pilot OTW data collection.  Our goal was to show a path towards achieving the sorts of per-class numbers that established deep-learning datasets like those in Table \ref{f:other_datasets}.  We were largely successful; we achieved over 200 examples of every VIRAT-relevant class except \textit{pulling}, \textit{riding} and cell-phone activities.

\begin{figure*}
\includegraphics[width=3.55in]{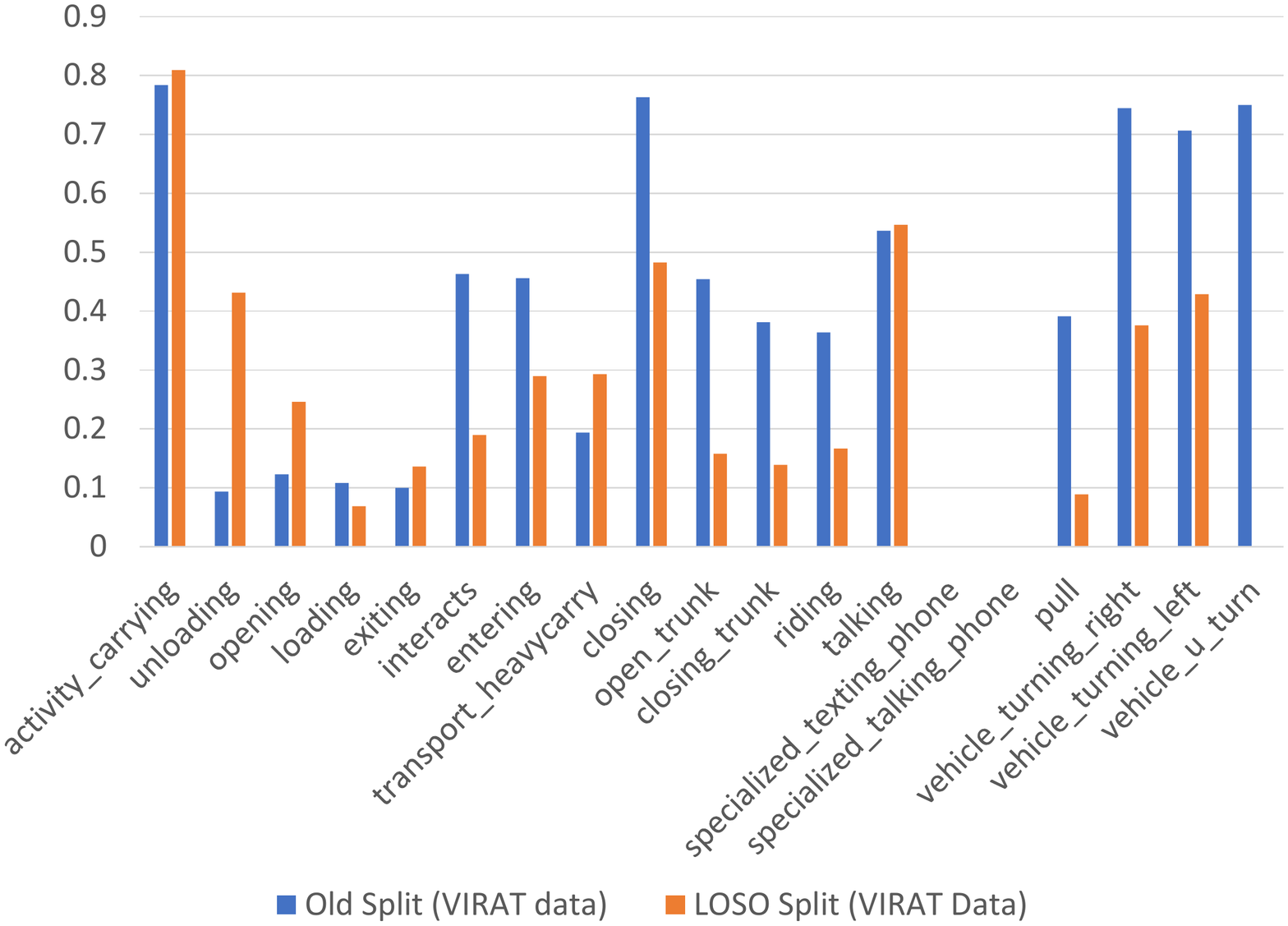}
\includegraphics[width=3.55in]{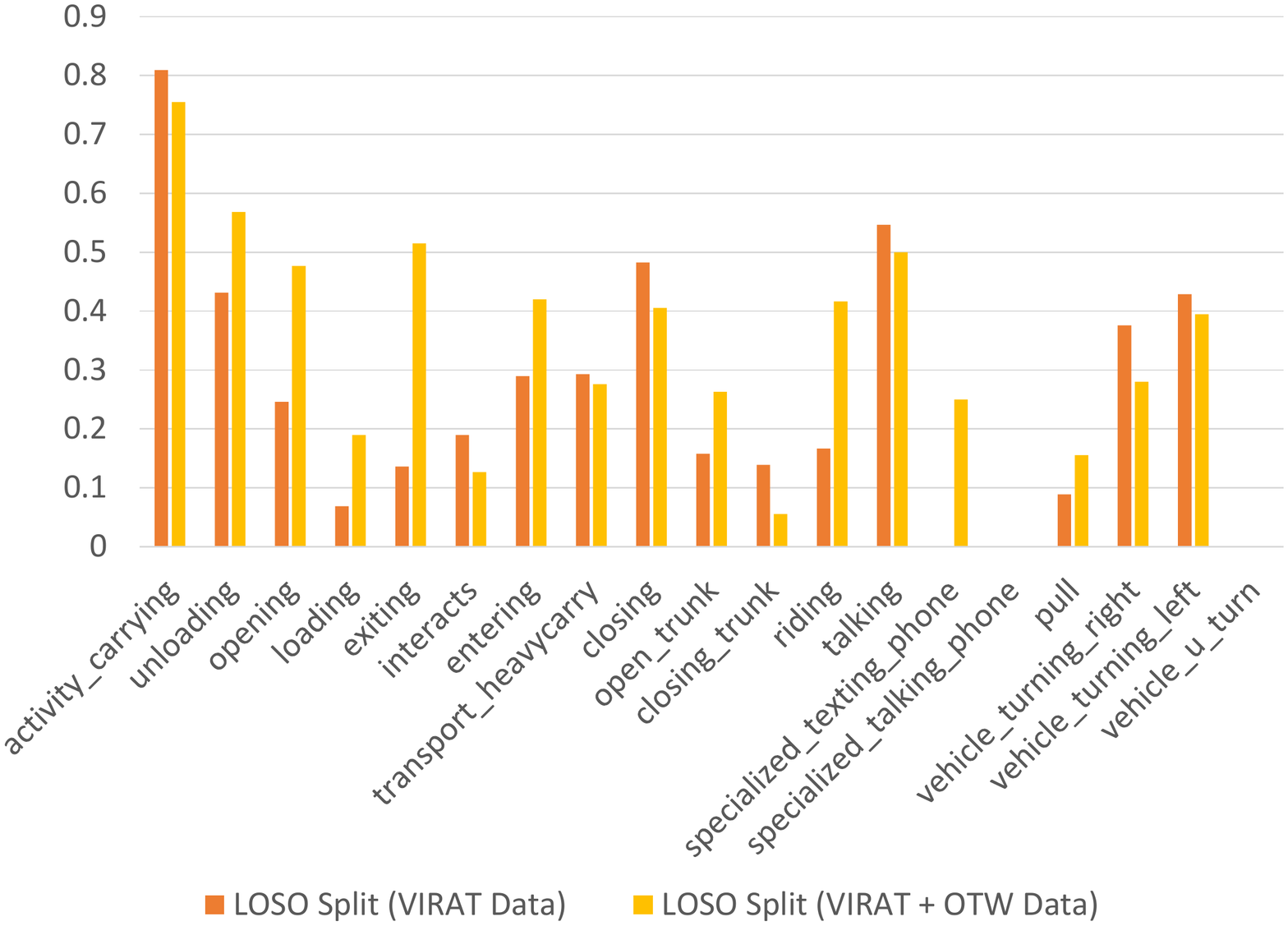}
\caption{(Left) Mean classification accuracy of a TSN trained and evaluated on the original VIRAT split, versus a TSN trained and tested using LOSO cross-validation.  (Right) Performance of TSNs trained on VIRAT and VIRAT+OTW using LOSO cross-validation.  This result demonstrates the primary benefit of the OTW dataset, to generalize activity representations to novel scenes.  
}
\label{f:loso}
\end{figure*}

\subsection{TSN Experiments on Default Split}
\label{ssec:tsn_default}

To begin to analyze the input of the OTW data, we applied it to the  train/validation split referenced above in Section \ref{ssec:phase1}.  The results are shown in Figure \ref{f:tsn_otwdiva_vs_diva}.

The improvements from the OTW dataset are consistent with expectations.  While the additional data does not provide much utility in improving vehicles activities (as it does not contain any examples of them), it pulls up the lower-performing classes significantly, offering from a 5 to a 40\% performance gain on these activities.  

We note that the default people-based activities labeled in the VIRAT dataset have 4 "pairs" of symmetric activities:  \textit{opening/closing}, \textit{entering/exiting}, \textit{loading/unloading}, \textit{open trunk/close trunk}.  As these behaviors look very similar except for the temporal ordering of the frames, they can often be confused.  We observe that this can result in a neural network effectively always guessing \textit{closing} for anything looking like \textit{opening} or \textit{closing}.  This results in a high score for \textit{closing} and a low score for \textit{opening}.  Similar phenomenology can be seen with high scores for \textit{entering} and low scores for \textit{exiting}.  We note that the tendency of the additional data from OTW is to pull up the low-scoring activities, rather than to dramatically improves the high scores.

\subsection{Leave One Scene Out (LOSO) Cross-Validation}
\label{ssec:loso}

%\begin{figure*}
%\includegraphics[width=7in]{figures/ttv_same.png}
%\caption{The DIVA Dataset contains the same activities performed by the same actors from the same %viewpoints in the Training, Validation and Test splits.
%}
%\label{f:ttv_same}
%\end{figure*}

Unfortunately, given that there are only a limited number of scenes in the VIRAT dataset, there is unfortunate coupling between the training and validation splits, as the same camera geometry and field of view are present in both.  The effect of this is that instead of learning the motions that represent the activities, networks can just overfit the context: learn where the left turn lanes are in a scene, learn where the parking spaces are.

To compensate for this, we introduce the concept of Leave One Scene Out (LOSO) cross-validation.  For each of the five scenes in VIRAT (\textit{0, 2, 400, 401, 500}), we declare it to be our validation set and train a model on the activities in the other four scenes.  The result is an independent model; the training data includes different scenes, viewpoints, and actors than the validation dataset.  We train 5 models, one using each scene as the left-out validation set, and sum the results, as shown in Figure \ref{f:loso}.

The results of doing a principled split of the data that does not allow for scene/viewpoint/actor bias is a dramatic decrease in performance of activities that are recognizable by context.  In particular, note that the vehicle turning activities drop by an average of over 30 points.  Activities that are also performed by only 1 or 2 actors also become harder to recognize; \textit{pull} activities in the dataset either involve a red radio-flyer cart (Scene 401) or a hand truck (Scene 0).  Without the ability to recognize those objects and declare that almost certainly a \textit{pull} event is occurring, performance drops there as well.  We also see drops across the other classes (notably excepting the \textit{unloading/loading} dyad); given that the same vehicles and people are used, we hypothesize that the network is taking advantage of these correlations.

Importantly, the addition of the OTW dataset in Figure \ref{f:loso} (Right) achieves its intended purpose of improved generalization.  On average, OTW offers a significant 8.3\% improvement in average classification accuracy on vehicle activities with humans.  However, on the more challenging activities (those with classification ac curacies of under 40\%), it offers an average improvement of 12.5\%.  This suggests that the OTW-style dataset offers a path forward to improving our accuracy in challenging classes.

% \subsection{Domain Mismatch}
% \label{ssec:domain_mismatch}

% Ultima

\section{Conclusions}
\label{sec:conclusions}

In this paper, we demonstrated that the addition of OTW data provided a significant improvement the performance of models on the VIRAT activity dataset.  In particular, when applied using Leave-One-Scene-Out cross-validation, it showed the ability to improve the generalization performance of our models.  However, there is still a significant, unexplained domain gap; models evaluated on the OTW dataset perform 15\% better, on average, than models evaluated using Leave-One-Scene-Out cross-validation on the VIRAT dataset.  That gap may be due to the peculiarities in the VIRAT dataset, far-field activities, camera quality, or other factors.  In future work, we plan to perform ablation studies to identify the causes of these mismatches, and expand the OTW dataset to address them.

The OTW dataset is subject to a Creative Commons Attribution 4.0 International license, and is available for download at \url{https://stresearch.github.io/otw}.

\medskip
\noindent {\bf Acknowledgement.}  Supported by the Intelligence Advanced Research Projects Activity (IARPA) via Department of Interior/ Interior Business Center (DOI/IBC) contract number D17PC00344. The U.S. Government is authorized to reproduce and distribute reprints for Governmental purposes notwithstanding any copyright annotation thereon. Disclaimer: The views and conclusions contained herein are those of the authors and should not be interpreted as necessarily representing the official policies or endorsements, either expressed or implied, of IARPA, DOI/IBC, or the U.S. Government.

{\small
\bibliographystyle{acm}
\bibliography{janus,jebyrne,gcastanon}
}

\end{document}